\pdfoutput=1

\documentclass[11pt]{article}

\usepackage[final]{acl}

\usepackage{times}
\usepackage{latexsym}

\usepackage[T1]{fontenc}

\usepackage[utf8]{inputenc}

\usepackage{microtype}

\usepackage{inconsolata}

\usepackage{graphicx}
\usepackage{xspace}
\usepackage{booktabs}
\usepackage{amsmath}
\usepackage{multirow}
\usepackage{arydshln}
\usepackage{subcaption}
\usepackage{xcolor}
\usepackage{array}     
\usepackage{makecell}
\usepackage{amsfonts}
\usepackage{soul}
\usepackage{pifont}
\usepackage{colortbl}
\usepackage{rotating} 
\usepackage{tabularx}
\usepackage{algorithmicx}
\usepackage{algpseudocode}
\usepackage{algorithm}

\usepackage{spverbatim}
\definecolor{lightgray}{gray}{0.95} %
\usepackage{fvextra}

\usepackage{fancyvrb}
\fvset{breaklines=true, breakanywhere=true}
\usepackage{fancybox}
\usepackage{float}

\usepackage{listings}
\usepackage{seqsplit}
\def \method{\textsc{VeriFact}\xspace}
\def \bench{\textsc{FactRBench}\xspace}

\title{\method: Enhancing Long-Form Factuality Evaluation \\with Refined Fact Extraction and Reference Facts}

\author{First Author \\
  Affiliation / Address line 1 \\
  Affiliation / Address line 2 \\
  Affiliation / Address line 3 \\
  \texttt{email@domain} \\\And
  Second Author \\
  Affiliation / Address line 1 \\
  Affiliation / Address line 2 \\
  Affiliation / Address line 3 \\
  \texttt{email@domain} \\}

\author{
 \textbf{Xin Liu},
 \textbf{Lechen Zhang}, 
 \textbf{Sheza Munir},
 \textbf{Yiyang Gu},
 \and \textbf{Lu Wang}
\\
Computer Science and Engineering\\
 University of Michigan\\
 Ann Arbor, MI
\\
    \texttt{\{\href{mailto:liuxincs@umich.edu}{liuxincs},
    \href{mailto:leczhang@umich.edu}{leczhang}, \href{mailto:shezamnr@umich.edu}{shezamnr}, \href{mailto:guyiyang@umich.edu}{guyiyang}, \href{mailto:wangluxy@umich.edu}{wangluxy}\}@umich.edu}
}

\definecolor{mygreen}{HTML}{74c476}
\definecolor{myred}{HTML}{f4a5a5}
\definecolor{myyellow}{HTML}{f8e19f}

\begin{document}
\maketitle
\begin{abstract}
Large language models (LLMs) excel at generating long-form responses, but evaluating their factuality remains challenging due to complex inter-sentence dependencies within the generated facts. 
Prior solutions predominantly follow a decompose-decontextualize-verify pipeline but often fail to capture essential context and miss key relational facts. 
In this paper, we introduce \method, a factuality evaluation framework designed to enhance fact extraction by identifying and \textit{resolving incomplete and missing facts} to support more accurate verification results. 
Moreover, we introduce \bench\footnote{Additional details about the project are available on its Hugging Face page: \url{https://huggingface.co/spaces/launch/factrbench}}, a benchmark that evaluates both precision and recall in long-form model responses, whereas prior work primarily focuses on precision. \bench provides \textit{reference fact sets} from advanced LLMs and human-written answers, enabling recall assessment. 
Empirical evaluations show that \method significantly enhances fact completeness and preserves complex facts with critical relational information, resulting in more accurate factuality evaluation. 
Benchmarking various open- and close-weight LLMs on \bench indicate that larger models within same model family improve precision and recall, but high precision does not always correlate with high recall, underscoring the importance of comprehensive factuality assessment.
\end{abstract}

\section{Introduction}

\begin{figure}
    \centering
    \includegraphics[width=1\linewidth]{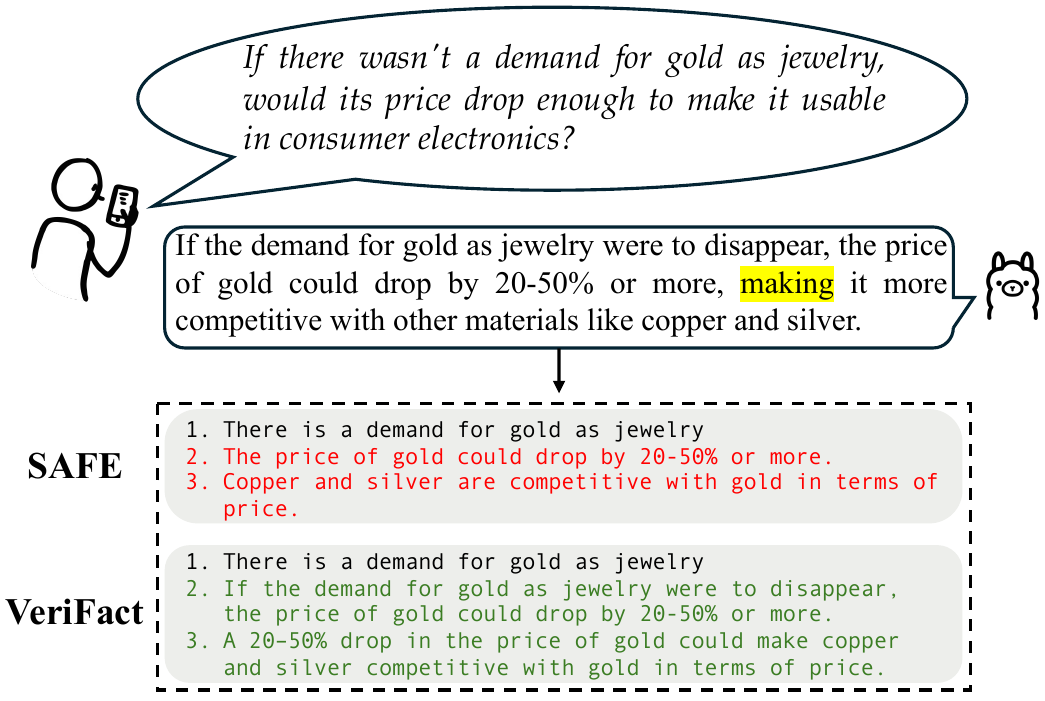}
    \caption{Sample facts extracted by SAFE~\cite{wei2024long} and our method \method from a model response. The highlighted token ``making'' indicates a causal relation \cite{prasad-etal-2008-penn} which is captured by \method but missed by SAFE. Facts marked in \textcolor{red}{red} are incomplete.
    }
    \label{fig:intro_case}
\end{figure}

Evaluating the factuality of long-form LLM responses is a challenging task, since the facts\footnote{In this paper, we use ``facts'' to refer to objective knowledge or claims that can be verified against external references.} in these outputs tend to be longer and more complicated, requiring a broader inter-sentence context for accurate assessment. 
Recent factuality evaluation methods~\cite{DBLP:conf/emnlp/MinKLLYKIZH23, Wang2023FactcheckGPTEF, wei2024long, DBLP:journals/corr/abs-2410-22257} all adopt a \textit{decompose}-\textit{decontextualize}-\textit{verify} pipeline approach. 
They first extract facts by decomposing a response and decontextualizing them (e.g., by resolving ambiguous mentions) into self-contained forms, and then independently verify each fact against external sources such as Wikipedia \cite{DBLP:conf/emnlp/MinKLLYKIZH23} or Google Search \cite{wei2024long}. 

There are two major issues with the existing methods. 
First, though effective for model responses with a simple discourse structure (e.g., a biography), they often fail to extract and verify facts covering longer text spans, which are frequently observed in long-form responses. 
As shown in Figure \ref{fig:intro_case}, when evaluating an answer to address a question about gold price, SAFE \cite{wei2024long}, a recent factuality evaluation method, extracts \textbf{incomplete facts}, such as ``The price of gold could drop by 20-50\% or more'' which loses essential context of ``if the demand for gold as jewelry were to disappear''. This omission weakens the verification step, leading to incorrect fact labels. 
Additionally, none of the facts extracted by SAFE captures the key \textbf{causal relation} (indicated by ``making''). 
Second, while most prior work emphasizes precision of factuality, few consider the \textbf{recall} of the generated answers, i.e., how comprehensively a response covers relevant facts to the query. Although some works like F1@K \cite{wei2024long} introduces a form of recall, it relies on a fixed and question-agnostic K that may misrepresent factual coverage, leaving recall of facts still under-studied.

To address these issues, our first contribution is a factuality evaluation method capable of handling complex inter-sentence dependencies in long-form text, built on the established decompose-decontextualize-verify pipeline. 
Our approach, \textbf{\method} (Verification of refined Facts), resolves the \textbf{incomplete facts} and \textbf{missing facts} that exists in the fact extraction component of prior methods. 
While previous work \cite{DBLP:conf/emnlp/GunjalD24} has attributed some incomplete facts to ambiguous entities, our human annotation study reveals that only 38\% of incomplete facts are due to ambiguity. The remaining ones can be categorized into \textit{missing comparandum} (where a comparison is implied but one item is absent) and \textit{omitted condition} (where a fact holds true only under a specific condition, as illustrated in Figure \ref{fig:intro_case}), and \textit{others}. 
\method addresses incomplete and missing facts by using multiple LLM judges to flag issues, which are then refined into self-contained facts that retain critical context and relations.
Finally, we verify the correctness of these refined facts by querying Google Search and comparing each fact against the retrieved evidence.

Our second contribution is \textbf{\bench}, a long-form generation benchmark that provides \textbf{reference facts} for assessing the recall of factuality in addition to precision. \bench consists of real-world queries sourced from (i) in-the-wild user prompts from FactBench \cite{DBLP:journals/corr/abs-2410-22257} and (ii) recent (post-January 2024) human-written knowledge-inquiring questions from Reddit (e.g., \texttt{r/askscience}). This design ensures relevance to real-world applications while reducing the risk of data leakage during LLM pretraining. 
To collect reference facts for queries from FactBench, where no ground-truth answer exists, we prompt four state-of-the-art LLMs, Claude3.5-Sonnet \cite{Claude_35}, GPT-4o \cite{DBLP:journals/corr/abs-2410-21276}, Gemini-1.5 \cite{DBLP:journals/corr/abs-2403-05530}, and Llama3.1-405B-instruct \cite{DBLP:journals/corr/abs-2407-21783}, to generate responses, ensuring broad knowledge coverage. 
We apply \method to extract and verify self-contained facts, retaining only those confirmed as correct for the reference set.
For Reddit-sourced queries with human-written answers, we extract facts from multiple highly-voted responses to serve as reference facts. 
Moreover, to ensure reproducibility, we release the web pages retrieved from Google Search as part of \bench's external knowledge resources for verification. To our knowledge, \bench is the first long-form factuality evaluation benchmark to include the \textbf{complete webpages}. This stable evidence set enables consistent evaluations by mitigating challenges from evolving online content and access limitations. 

Experimental results on fact extraction show that \method significantly reduces the extraction of incomplete facts by 19.2\% compared to the best comparison method. Additionally, the number of missing facts decreases by 37\% after applying \method's refinement stage, resulting in a more reliable factuality evaluation.
Furthermore, using \bench, we benchmark twelve frontier LLMs, including nine open-weight and three closed-weight models, with both precision and recall, and derive several key findings:
(\textbf{i}) Larger models within the same family generally achieve better precision and recall.
(\textbf{ii}) The models with the highest precision do not necessarily achieve the highest recall, emphasizing the need to consider both metrics for a more comprehensive factuality evaluation.
(\textbf{iii}) Closed-weight models tend to exhibit higher recall, while the largest open-weight models (e.g., Mistral-123B, Llama3.1-405B, Qwen2.5-72B) demonstrate highly competitive performance, particularly in precision, sometimes rivaling their closed-weight counterparts and showcasing significant progress in open model capabilities.

\section{Related Work}

Factuality evaluation has expanded beyond short-form QA \cite{lin2022truthfulqameasuringmodelsmimic, li2023haluevallargescalehallucinationevaluation, chen2023felmbenchmarkingfactualityevaluation} to long-form text, with benchmarks designed for more complex assessments. 
Existing benchmarks, such as FactScore \cite{DBLP:conf/emnlp/MinKLLYKIZH23}, LongFact \cite{wei2024long}, ExpertQA \cite{wei2024long}, FactCheck-Bench \cite{Wang2023FactcheckGPTEF}, assess factuality in long-form text but face domain limitations, model-generated prompts, or scalability issues.
Unlike previous work, \bench incorporates real-world hallucination-prone prompts from FactBench, and additional prompts from Reddit for greater topical variety and complexity.

Fact extraction and verification pipelines have emerged as prominent methodologies for long-form text factuality evaluation. 
Methods such as VERIFY \cite{DBLP:journals/corr/abs-2410-22257}, SAFE \cite{wei2024long}, and FactCheck-GPT \cite{Wang2023FactcheckGPTEF} adopt a decompose-decontextualize-verify paradigm, wherein LLM outputs are decomposed into atomic claims, refined for clarity, and subsequently verified against external sources. However, these approaches overwhelmingly focus on the precision of the facts,
while largely neglecting factual recall, i.e., whether a response sufficiently covers the relevant facts. 
Although SAFE
introduces a recall-oriented metric, F1@K, it relies on a fixed hyperparameter K across all prompts, which can overestimate coverage for simple questions or underestimate it for complex ones.
As such, factual recall remains an underexplored dimension in long-form evaluation. 
Unlike previous studies, \bench enhances factuality evaluation by including reference facts, allowing the computation of recall, and releasing full web documents to ensure reproducibility.
Moreover, recent research highlights challenges in fact extraction, particularly in decomposition and decontextualization \cite{DBLP:journals/corr/abs-2411-02400, DBLP:conf/emnlp/GunjalD24}. 
These methods mainly resolve entity ambiguity but fail to account for incomplete facts and missing key relational facts, which are critical for assessing factuality in long-form text \cite{chan2024chatgptevaluationsentencelevel}. Conversely, \method captures incomplete facts and reconstructs inter-sentence relations, enhancing factuality evaluation.

\section{An Annotation Study on Limitations of Fact Extraction}

\label{sec:limitations}
\subsection{Preliminaries}
Since existing factuality evaluation methods struggle with extracting facts that rely on inter-sentence dependencies, as illustrated in Figure \ref{fig:intro_case}, we first conduct an annotation study to identify the key issues. 
First, we randomly select 14 prompts\footnote{These 14 prompts are excluded from \bench.} from FactBench \cite{DBLP:journals/corr/abs-2410-22257}, a benchmark consisting of in-the-wild user prompts.
Then, we sample answers from two commonly used LLMs: GPT4o-mini and Llama3.1-8B, and apply the unit extractor of SAFE. By manually analyzing the resulting 451 extracted facts, we identify two main issues: \textbf{Incomplete Facts} and \textbf{Missing Facts}.

\textbf{Incomplete Facts} arise when extracted facts relies on additional context or other facts to be correctly interpreted. Based on our manual analysis, we categorize incomplete facts into four subtypes: \emph{ambiguous concept}, \emph{missing comparandum}, \emph{omitted condition}, and \emph{other}.
\textbf{Missing Facts} occur when extracted facts fail to capture inter-sentence relations that are critical for interpretation. Referring to the level-1 discourse relations defined in Penn Discourse TreeBank (PDTB) \cite{prasad-etal-2008-penn}, we adopt and focus on two subtypes: \texttt{temporal} and \texttt{contingency}, which we find to substantially impact the factuality of model-generated answers.
Detailed definitions and examples for each subtype are provided in Appendix \ref{apdx:detail_definition_incomplete_missing}.

\begin{figure*}[ht]
    \centering
    \includegraphics[width=1\linewidth]{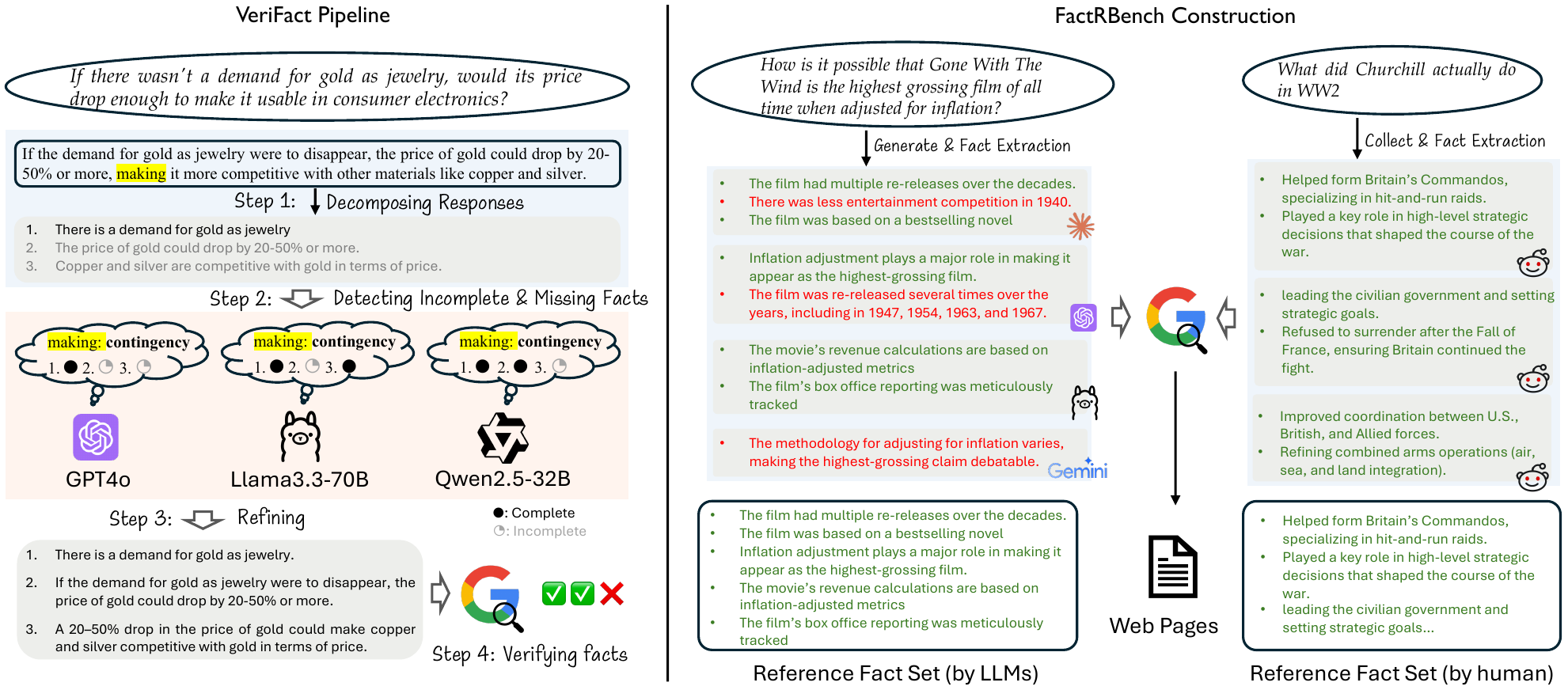}
    \caption{
    [Left] Our evaluation method, \method, employs multiple LLM judges to automatically identify incomplete facts (\textcolor{gray}{gray}) and missing facts from missing spans ({\sethlcolor{yellow}\hl{highlighted}}) in SAFE. It then applies reflection to correct these issues, ensuring the extracted facts are complete and capture all essential relations.  
    [Right] Overview of the process for collecting the reference fact set and web pages in \bench. 
    Sentences marked in \textcolor{red}{red} are contradicted or undecided, as verified by \method, and are excluded from the reference fact set.
}
    \label{fig:method}
\end{figure*}

\subsection{Human Annotation}
\label{sec:human_annotation}
Based on these observations, we employ four annotators, all fluent English speakers with relevant research experience in fact-checking, to identify incomplete and missing facts from all 451 extracted facts, with each fact independently annotated by all four. To facilitate annotating missing facts, we first apply a word-mapping algorithm\footnote{Details in Appendix~\ref{sec:mapping_alg}} to identify spans present in the original text but missing from the extracted facts. Each annotator then checks these ``missing spans'' to determine if they are missing facts or not. More details about the annotation guidelines, inter-annotator agreement, and the method we used to merge annotations are provided in Appendix \ref{apdx:preliminary_annotation_guideline}.

Our annotation reveals that 12.68\% of the facts extracted by SAFE are incomplete, with an average of 1.24 missing facts per response out of an average of 24.5 extracted facts. Among these incomplete facts, 38\% involve ambiguous entities, 6\% lack a comparandum, and 51\% omit a condition.
Since incomplete facts can alter the original meaning of an answer, they undermine the reliability of the evaluation. Meanwhile, missing facts may cause important relation errors to be overlooked.
Both issues significantly impact the overall trustworthiness of factual evaluation.

Based on our observation, incomplete facts often result from the decontextualization stage failing to preserve necessary context, while missing facts typically arise from incomplete decomposition. Rather than fixing these issues within each stage separately, we introduce an extra refinement step that detects and resolves both incomplete facts and missing facts. 

\section{\method: Factuality Evaluation of Long-form Responses}
In this section, we introduce \method, a pipeline evaluating factuality of long-form LLM responses in four steps. As illustrated in Figure \ref{fig:method}. \method builds upon the \textit{decompose}-\textit{decontextualize}-\textit{verify} pipeline of SAFE and introduces a new reflection-based decontextualization approach to address the issues of incomplete and missing facts in decomposition. In Step 1, we decompose LLM responses into facts by adopting the same method as SAFE. In Step 2, we use LLMs to identify incomplete and missing facts in the decomposed facts (\S\ref{sec:reflection}). Then, our Step 3 refines these detected issues through self-reflection (\S\ref{sec:self_refine}). Finally, in Step 4, we integrate the strengths of FactCheck-GPT and VERIFY to evaluate the correctness of refined facts based on Google Search (\S\ref{sec:fact_evaluation}).

\subsection{Detecting Incomplete and Missing Facts}
\label{sec:reflection}
To automate the detection of incomplete and missing facts, we employ three LLMs\footnote{These models were selected based on their agreement with human annotators, their coverage of human-identified problematic facts, and computational efficiency. Details can be found in Appendix~\ref{apdx:choice_of_llms}} (GPT-4o, Llama 3.3-70B, Qwen 2.5-32B) to replicate the annotation process followed by human annotators. Specifically, when identifying incomplete facts, we construct a prompt\footnote{The prompt can be seen in Appendix \ref{apdx:completeness_check}} that includes definitions for each subtype of incompleteness in \S\ref{sec:limitations} and instruct the LLM to determine whether a fact is incomplete, specify the exact subtype, and provide a rationale for its decision. When annotating missing facts, we follow the same process as human annotators by first using the word-mapping algorithm to identify potentially missing spans. We then construct a prompt\footnote{The prompt can be seen in Appendix \ref{apdx:missing_relation_prompt}} that includes definitions of temporal and contingency relations from PTDB and instruct the LLM to determine whether a fact is missing, specify the exact category, and provide a rationale for its decision. To reduce false positives, we conduct an additional verification step by prompting LLMs to ensure that the detected missing spans are not yet covered in other extracted facts.

We find that the annotation agreement between a single LLM and merged human annotations was relatively low, with an average Cohen’s Kappa of 0.33 over three models. However, since the primary goal of annotation is to facilitate fact refinement in the next step, false positives are not a major concern as they can still be refined and remain correct. Thus, rather than focusing on agreement, we emphasize recall, which measures how many human-labeled incomplete and missing facts are identified by LLMs. To improve recall, we adopt an ensemble learning approach, merging the annotation results from all three models by taking the union of their positive instances. This strategy significantly increases recall, achieving scores of 0.89 and 0.85 for completeness labeling and missing fact detection, respectively, indicating that over 80\% of incomplete and missing facts can be identified through automated labeling. 

\subsection{Resolving Incomplete and Missing Facts}
\label{sec:self_refine}
Using the annotations from previous steps, we refine the detected incomplete and missing facts respectively. (1) For incomplete facts, we prompt LLMs to revise them by incorporating missing context, ensuring they stand alone without additional dependencies. The prompt includes the original fact, its incompleteness category, and the explanation generated in the previous detection step. (2) For missing facts, we prompt LLMs to generate an additional fact that explicitly describes the overlooked temporal or contingency relation. The input includes the extracted fact, the missing relation type, and the corresponding missing phrase identified from previous steps. By combining these two refinement strategies, we ensure extracted facts are both self-contained and contextually complete, enabling more reliable factual evaluation. Relevant prompts are provided in Appendix \ref{apdx:refining}.

\subsection{Fact Evaluation}
\label{sec:fact_evaluation}
We follow the common workflow in the previous literature for fact evaluation, selecting FactCheck-GPT \cite{Wang2023FactcheckGPTEF} as our backbone, which was shown to have the highest precision in fact evaluation among various methods \cite{DBLP:journals/corr/abs-2410-22257, Wang2023FactcheckGPTEF, wei2024long}. Specifically, for each extracted fact, we generate multiple queries, call the Serper\footnote{https://serper.dev/} Google Search API, and extract snippets from the retrieved webpages as evidence. We then adopt the design improvements of VERIFY \cite{DBLP:journals/corr/abs-2410-22257}, using Llama 3.3-70B to classify the fact as \texttt{Supported}, \texttt{Contradicted}, or \texttt{Undecided} based on the gathered evidence. More details are available in Appendix~\ref{sec:fact_evaluation_details}.

\section{\bench}
While \method mitigates issues of incomplete and missing facts, thoroughly evaluating long-form responses demands more than just measuring correctness (i.e., \textit{precision}). We also need to assess coverage (i.e., \textit{recall}) to confirm that all essential facts are captured. Existing benchmarks typically emphasize verifying extracted facts for correctness alone, offering no robust way to gauge recall. 
To address this gap, we introduce \bench with real-world prompts (\S\ref{sec:prompt_source}) and construct \textbf{reference fact sets} to support the computing of recall (\S\ref{sec:collect_reference}). 
Finally, we explain how we compile supporting evidence for each prompt to facilitate fact verification of model responses (\S\ref{sec:compile_evidence}).

\subsection{Sourcing Real-World Prompts}
\label{sec:prompt_source}
To ensure alignment with real-world applications, we aim to collect prompts that reflect real user queries. The prompt sources in \bench consist of two main parts.
We first consider FactBench~\cite{DBLP:journals/corr/abs-2410-22257}, which contains prompts gathered from in-the-wild human-LLM interactions spanning 150 fine-grained topics. 
Since \bench aims to support the evaluation of recall, it requires reference facts and supporting evidence that comprehensively cover the knowledge relevant to each prompt.
However, some prompts in FactBench are highly divergent---these include open-ended questions that lack a specific answer and instead encourage broad, subjective reasoning, e.g. ``What are the top-10 restaurants in Tokyo?''
For such cases with many possible legit answers, it is infeasible to compile a complete set of knowledge context. To address this, we use GPT-4o to filter out these divergent questions.\footnote{Prompts for filtering divergent prompts in FactBench can be found in Appendix \ref{apdx:filtering_factbench_prompt}} With an additional human inspection step, this source results in a curated set of \textbf{649 prompts}.

The second part of our prompts is sourced from Reddit forums, specifically the \texttt{r/askscience}, \texttt{r/AskHistorians}, \texttt{r/AskEngineers}, and \texttt{r/AskEconomics} subreddits. 
These forums are moderated for evidence-based discussion, making the answers typically objective and thus well-suited for evaluating factual coverage. 
We collect prompts that appeared after Jan 2024, as the popular and frontier LLMs that are tested in this work have a knowledge cutoff before this period. This prevents the potential data leakage issue, making \bench a more robust benchmark for evaluating LLM factuality. 
To ensure the quality of human-written answers, we retain only prompts that have at least two responses, each has more than 35 upvotes and contains at least 70 words. Additionally, we filter out divergent prompts to ensure that the human-written answers provide comprehensive knowledge coverage relevant to the prompt. 
This gives us \textbf{447 prompts}.
In total, \bench contains \textbf{1096 prompts}.

\subsection{Constructing Reference Fact Sets}
\label{sec:collect_reference}
To support the evaluation of recall, we construct a reference fact set for each prompt, ensuring comprehensive coverage of relevant facts. Recall is then computed by measuring how many of these reference facts appear in the model-generated content. 
For Reddit-sourced prompts, we directly use the human-written responses, and apply \method to extract independent facts from them, forming the reference fact set.
By doing so, we get an average of \textbf{2.83} human responses per question and \textbf{64.17} reference facts per response.
Since FactBench prompts lack human-written answers, we adopt a multi-source approach to ensure a more comprehensive and unbiased reference. 
Specifically, we prompt four advanced LLMs: GPT-4o, Claude 3.5 Sonnet, Gemini 1.5, and Llama 3.1-405B-instruct, to generate responses for each prompt. We then apply \method to extract self-contained facts and assess their correctness.
We compile all \texttt{Supported} facts from the responses of four LLMs to construct the reference fact set, obtaining an average of \textbf{71.18} reference facts per response.

\begin{table}[t]
    \centering
    \renewcommand{\arraystretch}{1}
    \resizebox{0.48\textwidth}{!}{
    \begin{tabular}{lcccc}
        \toprule
        \makecell{\textbf{Benchmark}} & 
        \makecell{\textbf{Real-World}} & 
        \makecell{\textbf{Provide}\\\textbf{Evidence?}} & 
        \makecell{\textbf{Provide}\\\textbf{Reference}\\\textbf{Facts?}} & 
        \makecell{\textbf{\#}\\\textbf{Prompts}} \\
        \midrule
        \textbf{FELM}             & \ding{55} & \ding{55} & \ding{55} & 847 \\
        \textbf{ExpertQA}        & \ding{55} & \ding{55} & \ding{55} & 484 \\
        \textbf{FactScore}       & \ding{55} & \ding{55} & \ding{55} & 500 \\
        \textbf{LongFact}                  & \ding{55} & \ding{55} & \ding{55} & 2280 \\
        \textbf{FactCheckBench}          & mixed    & \ding{55} & \ding{55} & 94 \\
        \textbf{FactBench}     & \checkmark & \ding{55} & \ding{55} & 1000 \\
        \rowcolor{green!15}
        \textbf{\bench}                                     & \checkmark & \checkmark & \checkmark & 1096 \\
        \bottomrule
    \end{tabular}
    }
    \caption{Statistics of various factuality benchmarks. \bench is the first benchmark to evaluate recall while providing evidence for each prompt to facilitate fact-checking.
    }
    \label{tab:statistics}
\end{table}

\subsection{Collecting External Evidence} 
\label{sec:compile_evidence}
In addition to providing a reference fact set, we include supporting evidence for each prompt in \bench for fact verification. For prompts sourced from FactBench, we store complete webpages while evaluating the accuracy of advanced LLM responses. Similarly, for prompts from Reddit, we collect webpages by feeding human-written answers into \method for fact verification and storing all webpages retrieved via Google search. We list the statistics of these collected webpages in Appendix Table \ref{tab:crawl_statistics}.
By establishing a stable evidence set, \bench ensures reproducible evaluations, allowing researchers to assess their models consistently while mitigating challenges posed by evolving online sources and accessibility constraints.

Table~\ref{tab:statistics} compares \bench statistics with various long-form factuality benchmarks. Notably, \bench is the only benchmark that (\textbf{i}) provides full web pages fact verification and (\textbf{ii}) includes reference facts for recall evaluation, both are absent in all other benchmarks.

\section{Fact Extraction Evaluation}
We start by evaluating the fact extraction component of \method and the existing methods. We first describe the annotation process for collecting \textit{human-constructed reference facts} (\S\ref{sec:human_annotated_reference}), then introduce baselines (\S\ref{sec:baselines}) and discuss results (\S\ref{sec:extraction_eval_results}).

\subsection{Human-Annotated Extraction Facts}
\label{sec:human_annotated_reference}

To fairly compare \method and other fact extraction methods, we construct an evaluation set consisting of model response, human-annotated ground-truth extraction facts pairs.
This allow us to assess the model-extracted facts by comparing them against the ground-truth.
To improve annotation efficiency, we prioritize responses where evaluation methods frequently extract incomplete or missing relational facts. Using GPT-4o, we annotate responses from GPT-4o-mini and Llama 3.1-8B of prompts in \bench, then select those with a high occurrence of such issues based on SAFE-extracted facts.
A total of \textbf{1,168 facts} are annotated by four annotators\footnote{These annotators are same as those introduced in \S\ref{sec:human_annotation}}, each reviewing half the dataset, ensuring every fact gets two independent annotations
If a fact is incomplete, annotators revise it by adding necessary context to make it self-contained. 
Additionally, annotators check for any missing facts and manually add them if necessary.
This process produces two sets of human-written facts per response, both serving as reference fact sets. We evaluate model-extracted facts against each set separately and report the average results.
The annotation interface is provided in Appendix \ref{apdx:interface}.

\subsection{Fact Extraction Baselines}
\label{sec:baselines}
We evaluate \method against the fact extraction components of the following long-form factuality evaluation methods:
FactScore \cite{DBLP:conf/emnlp/MinKLLYKIZH23},
SAFE \cite{wei2024long},
FactCheck-GPT \cite{Wang2023FactcheckGPTEF},
and VERIFY \cite{DBLP:journals/corr/abs-2410-22257}.
Among these methods, FactScore lacks an explicit decontextualization stage, while the rest all follow a decompose-decontextualize-verify paradigm.
Details of these methods are listed in Appendix \ref{apdx:baselines}.

To further analyze the contributions of our selected LLM judges, we conduct additional ablation studies under three alternative configurations: (i) \textbf{Open models only} (Llama 3.3-70B + Qwen 2.5-32B), (ii) \textbf{Llama 3.3-70B only}, and (iii) \textbf{Qwen 2.5-32B only}.

\subsection{Metrics and Results}
\label{sec:extraction_eval_results}
Since human annotation may not fully cover all reference facts, we supplement recall evaluation to assess the \textbf{Coverage of Human-written Facts} (measured by GPT-4o \footnote{The prompt is provided in Appendix \ref{apdx:human_fact_recall}}) with additional metrics, including the ratio of incomplete facts (\textbf{Incomplete Facts (\%)}) and the average number of missing facts (\textbf{\# of Missing Facts}) identified by GPT4o, using the prompt introduced in \S\ref{sec:reflection}.

The evaluation results of \method and the comparison methods are shown in Table \ref{tab:fact_extraction}. The table indicates that \method generates the fewest incomplete facts and covers the most human-annotated reference facts, as evidenced by its lowest ratio of incomplete facts and highest recall among all methods.
This demonstrates the effectiveness of \method in extracting self-contained facts.
Additionally, \method reduces the average number of missing facts of SAFE by 37\%, i.e., from 1.22 to 0.76, further highlighting its ability to capture important relations. 
Interestingly, even the single-judge configuration (Qwen 2.5-32B only) yields competitive performance, achieving fewer incomplete facts and higher recall of human-written facts compared to all baselines. This underscores the robustness and practical applicability of our proposed method in simpler configurations.

We observe that other evaluation methods, such as FactScore and VERIFY, report a relatively lower number of missing facts. This is because the facts extracted by these methods are less fine-grained compared to others, resulting in fewer missing spans. This is reflected by the average number of extracted facts per response: 24.2 for FactScore, 18.5 for VERIFY, and 32.3 for \method, respectively. 
However, decomposing responses into fine-grained facts is crucial because it allows users to isolate specific claim components for targeted evidence retrieval and precise error localization \cite{DBLP:conf/emnlp/GunjalD24}.

\begin{table}[t]
    \centering
    \resizebox{0.48\textwidth}{!}{
    \begin{tabular}{lccc}
        \toprule
        \textbf{Method}
        & \begin{tabular}[l]{@{}l@{}}\textbf{Incomplete}\\ \textbf{Facts (\%) $\downarrow$}\end{tabular}
        & \begin{tabular}[l]{@{}l@{}}\textbf{Missing}\\ \textbf{Facts (\#) $\downarrow$}\end{tabular}
        & \begin{tabular}[l]{@{}l@{}}\textbf{Coverage of}\\ \textbf{Human Facts $\uparrow$}\end{tabular} \\
        \midrule
        FactScore \cite{DBLP:conf/emnlp/MinKLLYKIZH23}      & 70.4             & 0.86             & 60.0             \\
        SAFE \cite{wei2024long}           & 56.7             & 1.22             & 77.5             \\
        Factcheck-GPT \cite{Wang2023FactcheckGPTEF}  & 41.7 & 1.46             & 72.5             \\
        VERIFY \cite{DBLP:journals/corr/abs-2410-22257}         & 43.4             & \textbf{0.51}    & 78.7 \\
        \method       & \textbf{22.5}    & \underline{0.76} & \textbf{87.1}    \\
        ~~--~Open Models Only & \underline{26.3} & 0.76 & \underline{86.4} \\ 
        ~~--~LLaMA 3.3-70B Only & 30.6 & 0.82 & 83.5 \\ 
        ~~--~Qwen 2.5-32B Only & 26.7 & 0.78 & 85.4 \\
        \bottomrule
    \end{tabular}
    }
    \caption{
    Results of \method and compared methods on fact extraction using human-annotated reference facts. The best results are in \textbf{bold}, and the second-best results are \underline{underlined}. VERIFY extracts longer but fewer facts, with an average of 15.8 facts per response compared to 32.3 for \method. Consequently, VERIFY tends to miss fewer facts.
    }
    \label{tab:fact_extraction}
\end{table}

\begin{figure}[t]
    \centering
    \includegraphics[width=1\linewidth]{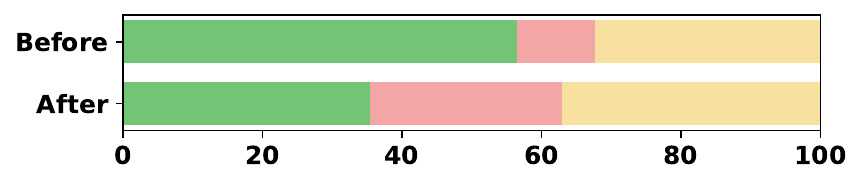}
    \caption{Ratio of correctness labels (\%) for facts whose labels changed before and after manual refinement (24.9\% refined facts have their correctness labels changed). We differentiate among {\sethlcolor{mygreen} \hl{\texttt{Supported}}}, {\sethlcolor{myred} \hl{\texttt{Contradicted}}}, and {\sethlcolor{myyellow} \hl{\texttt{Undecided}}}, as defined in \S\ref{sec:fact_evaluation}.
    }
    \label{fig:label_flipping}
\end{figure}

\begin{table*}[ht]
\centering
\resizebox{0.95\textwidth}{!}{
\begin{tabular}{lcccccccccccc}
\toprule
\textbf{Model} & \multicolumn{3}{c}{\textbf{Overall}} & \multicolumn{3}{c}{\textbf{FactBench}} & \multicolumn{3}{c}{\textbf{Reddit}} \\
& \textbf{Precision} & \textbf{Recall} & \textbf{F1} & \textbf{Precision} & \textbf{Recall} & \textbf{F1} & \textbf{Precision} & \textbf{Recall} & \textbf{F1} \\
\midrule
GPT4o            
& \textbf{83.30} & \textbf{52.42} & \textbf{64.35}
& 85.11 & \textbf{77.13}$^*$ & \textbf{80.92}
& \textbf{80.66} & 16.54 & 27.45 \\

Claude 3.5-Sonnet 
& 80.05 & 47.57 & 59.67
& 83.28 & 69.35$^*$ & 75.68
& 75.35 & 15.94 & 26.31 \\

Gemini 1.5-Flash  
& 83.02 & 49.01 & 61.63
& \textbf{85.45} & 70.71$^*$ & 77.38
& 79.48 & \textbf{17.50} & \textbf{28.68} \\

\hline
Mistral-7B        
& 74.91 & 36.00 & 48.63
& 77.79 & 51.96 & 62.30
& 70.72 & 12.82 & 21.71 \\

Mistral-24B
& 80.31 & 43.53 & 56.46
& 83.61 & 61.48 & 70.84
& 75.51 & \textbf{17.46} & \textbf{28.36} \\

Mistral-123B
& 82.24 & \textbf{46.60} & \textbf{59.49}
& \textbf{85.24} & 67.28 & \textbf{75.20}
& 77.88 & 16.57 & 27.33 \\

Llama3.1-8B       
& 63.02 & 37.33 & 46.89
& 68.27 & 54.28 & 60.48
& 55.40 & 12.73 & 20.70 \\

Llama3.1-70B      
& 71.90 & 40.23 & 51.59
& 73.40 & 58.00 & 64.80
& 69.72 & 14.42 & 23.90 \\

Llama3.1-405B     
& 76.51 & 46.11 & 57.54
& 78.80 & \textbf{68.40}$^*$ & 73.23
& 73.19 & 13.75 & 23.15 \\

Qwen2.5-8B        
& 75.09 & 40.25 & 52.39
& 76.09 & 58.66 & 66.25
& 73.64 & 13.53 & 22.86 \\

Qwen2.5-32B       
& \textbf{82.80} & 44.07 & 57.52
& 85.11 & 62.77 & 72.25
& \textbf{79.44} & 16.91 & 27.88 \\

Qwen2.5-72B       
& 82.14 & 44.61 & 57.82
& 84.97 & 64.12 & 73.09
& 78.02 & 16.29 & 26.95 \\

\bottomrule
\end{tabular}
}
\caption{
Evaluation results of various LLMs on \bench. We use the instruct version of all open-weights models. We highlight the best result in \textbf{bold} within open-weight models and the best result within closed-weight models separately. * indicates that the models' results may be inflated, as part of reference facts originate from their outputs. Models with highest precisions, i.e., Qwen2.5-32b, may not have the highest recall and F1, highlighting the importance of considering all metrics into factuality evaluation.
}
\label{tab:benchmark}
\end{table*}

\begin{figure*}[htbp]
    \centering
    \includegraphics[width=1.0\textwidth]{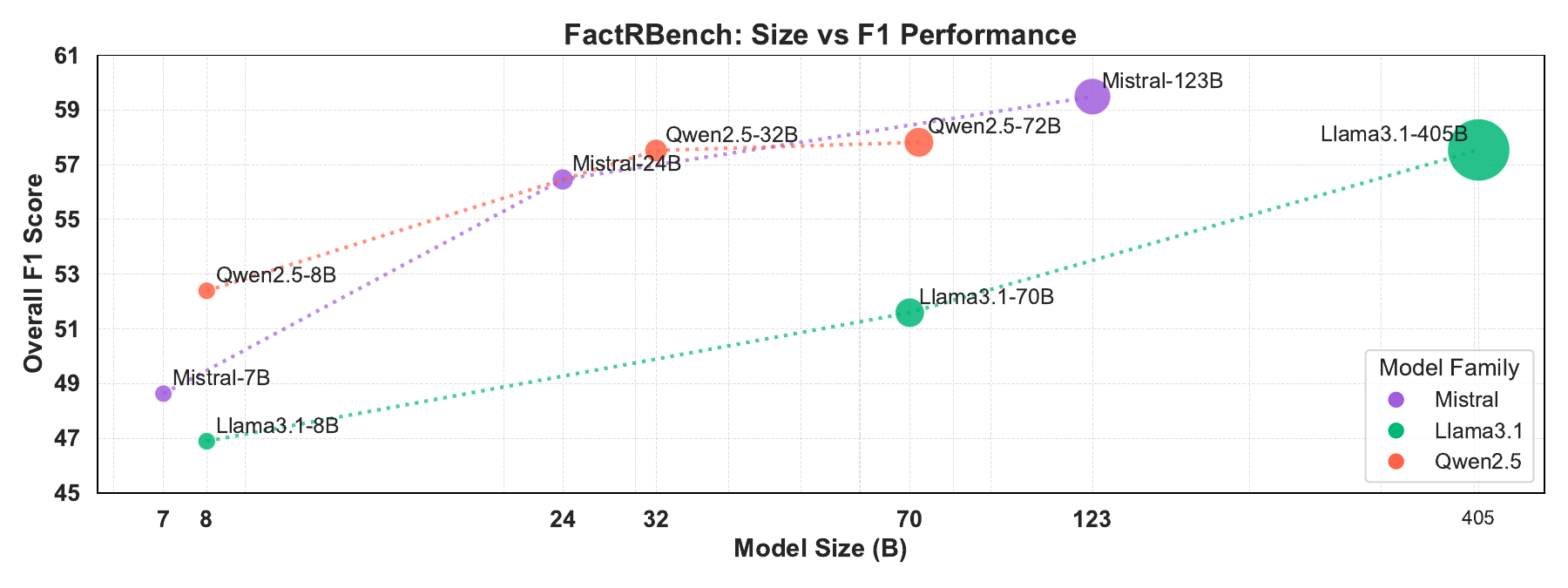}
    \caption{Visualization of the relationship between model size (billions of parameters) of open-weights models and overall F1 score on \bench. Larger markers indicate larger model sizes, and different colors represent different model families. Larger models within the same family generally achieve higher overall F1 scores.}
    \label{fig:size_vs_f1}
\end{figure*}

\subsection{Impact of Incomplete Facts on Factuality Evaluation}
Before benchmarking all factuality evaluation methods, we want to assess the extent to which incomplete facts may affect the evaluation pipeline. Therefore, we conduct an analysis on the correctness of incomplete facts before vs. after they are refined into complete ones.
We perform experiments using the human-annotated dataset introduced in \S\ref{sec:human_annotated_reference}. Specifically, we apply the fact verification process of \method to the extracted facts before and after manual refinement and evaluate how these modifications impact the factual correctness labels.

We find that 24.9\% of refined facts change their original correctness label from one label to a different one. Figure~\ref{fig:label_flipping} visualizes the distribution of correctness labels before and after refinement. The figure shows that the number of supported facts decreases sharply after refinement, while the number of contradicted facts increases significantly. 
Considering that previous studies assess factuality by counting the number of correct extracted facts, ignoring the issue of incomplete facts can lead to an overestimation of model factuality performance.

\section{Model Benchmarking with \bench}

\subsection{Experiment Settings}

\paragraph{Models.}
We evaluate twelve LLMs across five families.
For closed models, we assess GPT-4o, Gemini 1.5-Flash, and Claude 3.5-Sonnet. 
For open models, we consider models in the Mistral family with sizes of 7B \cite{DBLP:journals/corr/abs-2310-06825}, 24B (\texttt{Mistral-Small-3.1-24B-Instruct-2503}) \cite{mistral2025small3.1}, and 123B (\texttt{Mistral-Large-Instruct-2407}) \cite{mistral2024large2}, models in the Llama 3.1 family with sizes of 8B, 70B, and 405B, as well as the Qwen2.5 \cite{DBLP:journals/corr/abs-2412-15115} family at 8B, 32B and 72B. Detailed settings are available in Appendix \ref{apdx:model_setting}.

\paragraph{Evaluation Procedure.}
We apply \method to extract independent facts for the responses of all models, since it has the best fact extraction performance according to \S\ref{sec:extraction_eval_results}.
We follow the procedure outlined in \S\ref{sec:fact_evaluation} to evaluate the correctness of these extracted facts.
However, instead of querying web pages via Google Search, we directly use the pre-stored web pages associated with each response as the knowledge source.
The same knowledge source for all models ensures a consistent evaluation framework and enhances reproducibility.

We report three key factual evaluation metrics: 
(\textbf{i}) \textbf{Precision} measures the proportion of correct facts among all extracted facts.
(\textbf{ii}) \textbf{Recall} assesses the proportion of reference facts that are covered by the model response, and (\textbf{iii}) \textbf{F1} balances precision and recall in factual evaluation.
By employing these metrics, we ensure a rigorous and reproducible evaluation of precision and recall in LLM-generated content.\footnote{The details of these metrics are provided in Appendix \ref{apdx:metrics}.}

\subsection{Results and Analysis}
The results of twelve LLMs on \bench are shown in Table \ref{tab:benchmark}.
Looking first at open-weight models, we compare the overall results within the same model family, i.e., Mistral, Llama3.1, and Qwen2.5, we observe that both metrics improve as model size increases. This confirms that \emph{a larger model can generate not only more accurate but also more comprehensive answers}. {This trend is further illustrated in Figure \ref{fig:size_vs_f1}, which shows a clear correlation between model size and F1 score within each model family.

Comparing closed-weight and open-weight models reveals distinct trends. We observe that closed-weight models generally achieve higher recall on prompts from both sources. This aligns with the common observation that these models tend to provide more detailed answers to user queries. Among the evaluated closed models, GPT-4o achieves the highest overall recall. 
Nevertheless, the larger open-weight models, such as Mistral-123B, Llama3.1-405B, and Qwen2.5-72B, also demonstrate strong performance. Their capabilities, especially in terms of precision, sometimes rival those of closed-weight models, showcasing the rapid progress in open model development.
Within closed models, Gemini 1.5-Flash achieves the highest precision in in-the-wild prompts sourced from FactBench, however, it exhibits lower recall and F1 scores compared to GPT-4o. This emphasizes the need to consider multiple metrics when evaluating factuality in long-form responses.

Comparing results from the two prompt sources, we find that all models perform worse on Reddit-sourced prompts, with both precision and recall decreasing. We believe this is because human-written answers, which serve as the reference fact set on Reddit, are far more diverse than LLM-generated responses. Evaluating model outputs against such diverse references presents a significant challenge in assessing the coverage of their responses.

\begin{table}
    \centering
    \small
    \begin{tabularx}{\linewidth}{X}
    \toprule
    \textbf{Question:} Why is English not the official language of the United States even though every official document is written in English? \\
    \textbf{Response:} Although English is the most widely spoken language and is often considered the de facto language of the United States (meaning it is the most common language used in everyday life and government), the U.S. does not have a federally mandated official language, including at the Federal Level. \\ 
    \textbf{Model:} Llama3.1-70B \\\hline
    \textbf{Fact Reference Set:} \\
    1. The U.S. is a nation of immigrants, and enforcing an official language could be seen as exclusionary. \\
    2. Political resistance exists due to concerns about discrimination and civil rights. \\
    ...
    \\\hline
    \textbf{Precision:} 100.00\\
    \textbf{Recall:} 33.72\\
    \bottomrule
    \end{tabularx}
    \caption{
    An example illustrating the evaluation of a model response considering both precision and recall. The response achieves high precision but low recall, as it provides a correct but uninformative answer.
    \vspace{-6mm}
    }
    \label{tab:case_study}
\end{table}

\subsection{Case Study}

Table \ref{tab:case_study} presents an example of a response generated by Llama3.1-70B. While the response correctly states that English is not the official language of the United States, it does not provide any explanation as to why this is the case, which is considered more useful if given. As a result, it achieves perfect precision but remains uninformative, offering little value to the user. When compared to the fact reference set in \bench, it becomes clear that many important details are missing, leading to a low recall.
This demonstrates the limitation of relying solely on precision for evaluation and highlights the importance of recall.

\section{Conclusion}
In this work, we introduced \method, a factuality evaluation framework that improves fact extraction by resolving incomplete and missing facts to support accurate verification results. Additionally, we designed \bench, the first benchmark to assess both precision and recall, ensuring a more comprehensive evaluation of factuality. Empirical results show that \method effectively enhances fact completeness and preserves complex facts with critical relational information.
Findings from \bench underscore the importance of balancing accuracy and completeness in LLM development, guiding future advancements in the field.

\section*{Acknowledgements}
This work is supported in part by National Science Foundation through grant 2046016. 
We thank our colleague Farima Fatahi Bayat for insightful discussions during the early stages of this work. 
We thank Serper API (\url{https://serper.dev}) for generously providing access to their Google Search API, which enabled large-scale fact verification in our experiments.
We thank anonymous reviewers of ARR for their thoughtful comments. 

\section*{Limitations}
While \method and \bench introduce significant advancements in factuality evaluation for long-form responses, several limitations remain. 
(1) Although \bench provides a benchmark for both precision and recall, the recall metric is inherently dependent on the completeness of the reference fact set. While we mitigate this by including multiple LLM-generated and human-written reference facts, there remains a risk that some relevant facts are omitted, leading to underestimated recall scores.
(2) Despite improvements in handling incomplete and missing facts, the fact extraction process still depends on LLM-based annotation, which introduces possible biases. The ensemble approach helps improve coverage, but the agreement between human annotators and LLMs remains imperfect.
(3) The pipeline involves multiple LLM passes for fact detection, refinement, and verification, making it computationally expensive. Running \method at scale may not be feasible for real-time applications due to latency and resource constraints.

Future work can address these limitations by exploring more efficient fact-checking mechanisms, and expanding reference fact sets with human-constructed data.

\bibliography{acl_latex}

\appendix

\section{Appendix}
\label{sec:appendix}
\subsection{Fact Evaluation Details}
\label{sec:fact_evaluation_details}
Our fact evaluation procedure follows the methodology outlined in Factcheck-GPT \cite{Wang2023FactcheckGPTEF}, incorporating design improvements from VERIFY \cite{DBLP:journals/corr/abs-2410-22257}. We use Llama 3.3-70B-Instruct as the backbone LLM when needed.

Given an extracted fact, we first generate multiple search queries using the prompt-based paraphrasing technique in FactCheck-GPT for two rounds to maximize the likelihood of retrieving relevant evidence while avoiding excessive specificity or bias. We then use the Serper Google Search API to retrieve and scrape the webpages based on the generated queries. Next, we extract five snippets by calculating and ranking the relevance score between the query and all scraped paragraphs using the Cross-Encoder pipeline from FactCheck-GPT. Finally, we use the fact evaluation prompt from VERIFY, combined with the snippets extracted by FactCheck-GPT, to classify each fact into one of three categories: \texttt{Supported}, \texttt{Contradicted}, \texttt{Undecided}.

\subsection{Detailed Definition of Incomplete and Missing Facts}
\label{apdx:detail_definition_incomplete_missing}
\paragraph{Incomplete Facts.} 
    Incomplete facts can arise from several sources. One such source is \emph{ambiguous concept}, where the meaning depends on specifics that are left unstated. While prior work \cite{DBLP:conf/emnlp/GunjalD24} highlights ambiguous entities as a primary cause of incompleteness, 
    we find that ambiguity accounts for only a small portion of all incomplete facts, indicating the need to consider other types as well.
    Another cause of incompleteness is the \emph{missing comparandum}, where a comparative statement omits one item (e.g., ``The Intel i7 4770k is an older processor'' without specifying the reference point). We also identify \emph{omitted condition}, where the validity of a fact should depend on a hypothetical scenario, condition, or quoted reference but is left out, such as the example in Figure \ref{fig:intro_case}. Finally, we include an \emph{other} category for incomplete facts that do not fit these previous classifications.
    
    \paragraph{Missing Facts.}
    A key issue we address in this paper, which has not been tackled by previous work, is the failure of extracting facts that capture crucial \emph{inter-sentence relations}. Drawing on the Penn Discourse TreeBank (PDTB) \cite{prasad-etal-2008-penn}, we focus on two level-1 discourse relations, i.e., \texttt{temporal} and \texttt{contingency}, which often substantially influence the factuality of a model's answer.\footnote{We initially considered all four PDTB level-1 relations but found that important facts primarily fall within these two categories.}
    Temporal relations describe how arguments are related in time, clarifying whether events occur sequentially or simultaneously. Contingency relations, on the other hand, demonstrate how one argument provides a reason, explanation, or justification for the situation described by another argument.
    Consider the example in Figure \ref{fig:intro_case}: the verb ``making'' signals a contingency relation, but the facts extracted by SAFE fail to capture this relation. As a result, the extracted facts miss an essential piece of information, leading to an incomplete understanding of the event.

\subsection{Preliminary Annotation Guideline and Analysis}
\label{apdx:preliminary_annotation_guideline}
To study the issue of incomplete facts, annotators are asked to determine whether a fact is incomplete and, if so, specify the subcategory of incompleteness, as we find that asking for specific subcategories improves inter-annotator agreement.
To better integrate annotations from the four annotators, we also require each annotator to indicate their confidence level (high or low) for each label. We can thus merge the annotations as follows: a fact is classified as incomplete if any annotator marks it as such with high confidence or at least two annotators label it as incomplete. This mitigates the impact of annotators who might overlook certain incomplete facts. In terms of Cohen's Kappa, the average agreement between each annotator and the merged result is 0.73.

Annotators are also required to identify missing facts.
We observe that important relational fact are often missing in extracted facts when keywords or phrases (e.g. ``Afterward'', ``due to'') signaling temporal or contingency relations are omitted by LLM.
Based on this observation, we first apply a word-mapping algorithm (Details in Appendix~\ref{sec:mapping_alg}) to identify spans present in the original text but missing from the extracted facts. Each annotator then checks these ``missing spans'' to determine if they indicate a temporal or contingency relation. If so, the annotator marks it as a missing fact.
We collect four independent annotations for each fact and merge them by majority vote. The average agreement between each annotator and the merged results is 0.90, with a Cohen's Kappa of 0.69.

\subsection{Word Mapping Algorithm}
\label{sec:mapping_alg}
Algorithm \ref{alg:word_mapping} illustrates our word mapping algorithm. It identifies the missing spans for each fact by applying the longest common substring algorithm, and returns the spans where no fact is mentioned.

\begin{algorithm}[h]
\caption{Word Mapping Algorithm}
\begin{algorithmic}[1]
\State \textbf{Input:} Response text \(R\) and extracted facts \(\mathcal{F}=\{f_1, f_2, \dots, f_n\}\) (each possibly multi-line)
\State \textbf{Output:} Missing spans \(M\)
\State Initialize a binary indicator array \(I\) of length \(|R|\) with zeros.
\State Initialize empty lists \(M\) and \(RR\) (to store missing spans and their index ranges).
\For{each fact \(f \in \mathcal{F}\)}
         \State \((s, start, end) \gets \text{LongestCommonSubstring}(R, f)\)
         \If{\(s \neq \emptyset\)}
             \State Mark indices \(I[start:end]\) as 1.
         \EndIf
\EndFor
\State Identify contiguous segments in \(I\) with value 0 (i.e., record their start and end indices) and add the corresponding substrings of \(R\) to \(M\) while storing the ranges in \(RR\).
\State \Return \(M\)
\end{algorithmic}
\label{alg:word_mapping}
\end{algorithm}

\subsection{Choice of LLMs for Automatic Annotation}
\label{apdx:choice_of_llms}
In our pilot study, we explore various API-based LLMs (GPT-4o, Claude 3.5-Sonnet, Gemini 1.5-Flash) and open-source LLMs (DeepSeek-R1-32B, LLaMA 3.3-70B, Qwen 2.5-32B, and Qwen 2.5-70B) for automatically detecting incomplete and missing facts.

We compare their annotations of fact completeness and missing facts with the merged human annotations, reporting Cohen's Kappa and recall for identifying human-labeled incomplete and missing facts in Table~\ref{tab:agreements}. Based on agreement, recall, computational efficiency, and general applicability, we select GPT-4o, LLaMA 3.3-70B, and Qwen 2.5-32B for automatic labeling. Although GPT-4o demonstrates relatively lower recall compared to LLaMA 3.3-70B and Qwen 2.5-32B, we include it due to its widespread adoption and general usage, ensuring broader practical relevance. Furthermore, we intentionally select models from three distinct model families (GPT, LLaMA, and Qwen) to maximize diversity and generalizability, minimizing potential biases arising from any single architecture. Additionally, despite Qwen 2.5-70B achieving the highest recall among open-source models, we choose the smaller Qwen 2.5-32B model as it provides competitive performance at significantly lower computational costs, thereby improving the accessibility and practical efficiency of our approach.

\subsection{Baseline Description}
\label{apdx:baselines}
We use GPT-3.5-Turbo-0301 \cite{DBLP:conf/nips/BrownMRSKDNSSAA20} as the foundational model and apply the prompts from their paper for all baselines in decomposition and decontextualization.
\begin{itemize}
    \item \textbf{FactScore} \cite{DBLP:conf/emnlp/MinKLLYKIZH23}: FactScore measures the factual accuracy by deconstructing responses into individual factual components. It then determines the proportion of these components that can be verified using Wikipedia articles. FactScore represents one of the earliest approaches to fact extraction. It operates by feeding each sentence in a response individually into an LLM to extract atomic facts. However, it overlooks inter-sentential dependencies such as pronoun resolution, resulting in a high number of incomplete facts. Notably, it does not include an explicit decontextualization stage.
    \item \textbf{Search-Augmented Factuality Evaluator (SAFE)} \cite{wei2024long}: SAFE assesses the factual accuracy of long-form text by decomposing each sentence from the model response into atomic facts. It introduces a decontextualization stage to add relevant context, making each fact as self-contained as possible. This step reduces incomplete facts significantly. However, it can also introduce errors by failing to preserve the original meaning, which leads to an increase in missing facts.
    \item \textbf{Factcheck-GPT} \cite{Wang2023FactcheckGPTEF}: Factcheck-GPT is a framework designed to detect and correct hallucinations in model responses. It evaluates factuality using a unified pipeline: the entire response is input at once, and the model outputs a list of self-contained facts. This bypasses decomposition and decontextualization as separate steps. While this approach helps generate largely self-contained facts, it demands strong factual reasoning from the model, often resulting in many crucial facts being omitted and a high rate of missing facts.
    \item \textbf{VERIFY} \cite{DBLP:journals/corr/abs-2410-22257}: VERIFY is a structured pipeline for assessing the factuality of LMs. It includes both decomposition and decontextualization stages. Unlike SAFE, VERIFY directly decomposes the entire response into facts, producing less granular but more coherent outputs. It improves upon previous methods by introducing carefully crafted prompts with explicit and structured reasoning, substantially enhancing the quality of the decontextualization step.
\end{itemize}

\subsection{Detailed Model Settings}
\label{apdx:model_setting}
Since greedy search can cause some models to produce repetitive outputs, we generate responses from all LLMs using a temperature of 1.0.
For GPT-4o, we use the model version of 2024-05-13.
All open-weight models are run in float16 to reduce memory usage.

\subsection{Metrics of \bench}
\label{apdx:metrics}
\textbf{Precision} measures the proportion of correct facts among all extracted facts:
\begin{align}
    \mathbb{E}_{y\in \mathcal{Y}}(\frac{1}{|\mathcal{E}_y|} \sum_{e\in\mathcal{E}_y} \mathbb{I}[e \text{ is supported by } K]),
\end{align}
where $\mathcal{E}_y$ represents the extracted facts from the model response $y$, and $K$ is the knowledge source.
\textbf{Recall} assesses the proportion of reference facts that are covered by the model response:
\begin{align}
    \mathbb{E}_{y\in \mathcal{Y}} ( \frac{1}{|\mathcal{\bar{E}}_x|} \sum_{\bar{e} \in \mathcal{\bar{E}}_x} \mathbb{I}[\bar{e} \text{ is supported by } y] ),
\end{align}
where $\mathcal{\bar{E}}_x$ is the reference fact set for the prompt $x$, and $y$ is the model-generated response to $x$. 
Here we use the LLama3.3-70B-Instruct model to determine the entailment relation. We list the prompt in Appendix \ref{apdx:factual_recall}.

\subsection{Crawled Website Statistics}
We report the statistics of crawled website in Table \ref{tab:crawl_statistics}.
\begin{table}[h]
    \centering
    \resizebox{0.95\linewidth}{!}{
    \begin{tabular}{l c c c}
        \toprule
        \textbf{Response Source} & \textbf{Avg. Claims} & \textbf{Avg. Ref. Claims} & \textbf{Avg. Crawled Websites}\\
        \midrule
        Claude 3.5-Sonnet & 78.55 & 65.56 & 1729.55 \\
        Gemini 1.5-Flash & 83.83 & 71.63 & 1902.20 \\
        GPT-4o & 84.28 & 71.73 & 1866.74 \\
        Llama 3.1-405B & 99.40 & 77.23 & 2154.38 \\
        \midrule
        Reddit & 63.64 & 44.03 & 1537.46 \\
        \bottomrule
    \end{tabular}
    }
    \caption{Average number of claims, reference claims, and crawled websites per response in each source of response.}
    \label{tab:crawl_statistics}
\end{table}

\subsection{Knowledge Coverage of the Provided Webpages}

We provide webpages in \bench for fact correctness evaluation, serving as an alternative source of knowledge to Google Search.
To assess the knowledge coverage of these webpages, we compare evaluation results for the same set of extracted facts using both the provided webpages and Google Search.
Specifically, we randomly select 200 responses (100 from FactBench and 100 from Reddit) generated by Qwen2.5-32B and Llama3.1-70B. After applying \method, we obtain 10,751 facts from FactBench and 11,831 from Reddit, which are then evaluated using both the provided webpages (\textbf{offline}) and Google Search (\textbf{online}). The results are shown in Table \ref{tab:comparison_by_evidence_type}.

From the table, we observe that evaluations conducted using offline evidence exhibit a similar trend in the ratio of supported claims as those conducted using online evidence across both FactBench and Reddit. This demonstrates that the provided webpages serve as a reliable alternative source of evidence to online search engines.
However, the ratio of undecided labels is consistently higher with offline evidence, indicating that their knowledge coverage is not as comprehensive as that of online search results.
Therefore, while the provided webpages serve as a strong and practical alternative, particularly in scenarios where online access is limited, we recommend using an online search engine for evaluations requiring the highest precision and completeness.

\begin{table*}[h!]
\centering
\begin{tabular}{l l c c c c}
\toprule
\textbf{Source} & \textbf{Model Name} & \textbf{Evidence Type} & \textbf{Contradicted} & \textbf{Supported} & \textbf{Undecided} \\
\hline
\multirow{4}{*}{FactBench} 
    & Llama3.1-70b & offline & 0.05 & 0.68 & 0.28 \\
    & Llama3.1-70b & online  & 0.08 & 0.78 & 0.14 \\
    & Qwen2.5-32b  & offline & 0.03 & 0.60 & 0.38 \\
    & Qwen2.5-32b  & online  & 0.07 & 0.76 & 0.18 \\
\hline
\multirow{4}{*}{Reddit} 
    & Llama3.1-70b & offline & 0.03 & 0.56 & 0.41 \\
    & Llama3.1-70b & online  & 0.06 & 0.72 & 0.22 \\
    & Qwen2.5-32b  & offline & 0.03 & 0.69 & 0.28 \\
    & Qwen2.5-32b  & online  & 0.04 & 0.81 & 0.15 \\
\bottomrule
\end{tabular}
\caption{Model performance comparison by evidence type (offline vs. online) and data source (FactBench and Reddit). We report the ratio of extracted facts labeled as contradicted, supported, or undecided.}
\label{tab:comparison_by_evidence_type}
\end{table*}

\begin{table}
    \centering
    \resizebox{0.95\linewidth}{!}{
    \begin{tabular}{lcc}
    \toprule
        \textbf{Model} & \textbf{Cohen's Kappa} & \textbf{Recall} \\ \hline
        GPT4o & 0.336 & 0.625 \\
        Claude 3.5-Sonnet & 0.204 & 0.656 \\
        Gemini 1.5-Flash & 0.118 & 0.521 \\ \hline
        DeepSeek-R1-32B & 0.328 & 0.344 \\
        Llama 3.3-70B & \textbf{0.478} & 0.688 \\
        Qwen 2.5-32B & 0.327 & 0.750 \\
        Qwen 2.5-70B & 0.324 & \textbf{0.844} \\
        \bottomrule
    \end{tabular}
    }
    \caption{Evaluation of LLMs on fact completeness and missing fact annotations. Best results are highlighted in \textbf{bold}.}
    \label{tab:agreements}
\end{table}

\subsection{Prompts}

\subsubsection{Completeness Check}
\label{apdx:completeness_check}
{\tiny\begin{lstlisting}
# Task

Given a context and a claim extracted from the context, determine whether the claim is Dependent or Independent of the context.
* Independent: If the claim itself precisely reflects the original meaning of the context without further explanation.
* Dependent: If the claim requires additional context or detail to precisely reflect its original meaning.

A claim is Dependent if it requires additional context to reflect the original meaning. Categorize it into one of three types:

* Ambiguous Concepts/Pronouns
    The claim contains vague terms (e.g., "this method," "they") or pronouns lacking clear referents from the context.
    Example:
    Context: "Decarbonizing aviation requires SAFs."
    Claim: "They reduce emissions." ->Dependent (Ambiguous pronoun "they").
* Missing Comparison
    The claim implies a comparison (e.g., "more," "better") but omits the explicit comparison target stated in the context.
    Example:
    Context: "SAFs reduce emissions by 80% compared to jet fuels."
    Claim: "SAFs reduce emissions by 80%." -> Dependent (Missing "compared to jet fuels").
* Lack of Condition/Sources
    The claim omits critical contextual details, such as:
    - Temporal conditions (e.g., "as of 2023").
    - Hypothetical scenarios (e.g., "if regulations are adopted").
    - Sources/References (e.g., "According to the ICAO...").
    Example:
    Context: "As of 2023, the U.S. top 1% net worth is ~$10M (Smith et al., 2023)."
    Claim: "The U.S. top 1% net worth is ~$10M." -> Dependent (Missing both time and source).
# Example

Context:
"Fine-tuning in the context of deep learning refers to the process of taking a pre-trained model-typically one that has been trained on a large dataset-and making small adjustments to its weights and parameters to adapt it for a specific task or dataset. This approach leverages the knowledge the model has already acquired, allowing it to achieve better performance on the new task with less data and training time compared to training a model from scratch. Fine-tuning usually involves freezing some of the earlier layers of the model, which capture general features, while allowing the later layers to be retrained to capture task-specific features. "

Claim:
"Training a model being trained from scratch requires more data."

Your Response:
* Explanation: "The claim states that training a model from scratch requires more data, but it does not specify the comparison target for "more than".
* Classification: Dependent
* Dependent Type: Missing Comparison

[7 more demonstrations]

# Your Task

Context:
"[Insert the relevant context here]"

Claim:
"[Insert the extracted claim here]"

Your Response:
\end{lstlisting}
}

\subsubsection{Missing Relation Check}
\label{apdx:missing_relation_prompt}
Missing relation detection:
{\tiny\begin{lstlisting}
The PDTB annotation manual (\S 4.2: Sense Classification) defines various discourse relations:
  1.  Temporal Relations - Situations related by time.
  *  Synchronous: Events overlap in time.
  *  (E.g., The company operates under Chapter 11, giving it court protection while restructuring.)
  2.  Contingency Relations - Cause-effect relationships.
  *  Cause.Result: Arg1 provides a reason, Arg2 its effect.
  *  (E.g., The debt is declared equity, so it isn't deductible.)
  *  Cause.NegResult: Arg1 prevents the effect in Arg2.
  *  (E.g., Investors acted too late to avoid losses.)
  *  Cause+Belief: Evidence supports a claim.
  *  (E.g., Southern African nations manage elephants well, so their herds thrive.)
  *  Cause+SpeechAct: Reason for a speech act.
  *  (E.g., "Maybe I'm stuffy, but I wouldn't sell them," sniffed Bob.)
  *  Purpose: An agent takes action to achieve a goal.
  *  (E.g., A company sells radio stations to focus on programming.)
  *  Condition: One situation depends on another being realized.
  *  (E.g., If Congress agrees, the government will relinquish its stake.)
  *  Negative Condition: The effect occurs unless a condition is met.
  *  (E.g., Profits may remain low unless the Fed eases rates.)
  3.  Comparison Relations - Highlights differences or similarities.
  *  Contrast: At least two explicit differences between arguments.
  *  (E.g., Gold thrives in inflation; utility stocks thrive in disinflation.)
  *  Concession: A causal expectation is denied.
  *  (E.g., The plan worked, even though it didn't prevent the plunge.)
  *  Similarity: Highlights commonalities.
  *  (E.g., Just as 1980s markets transformed finance, so will the 1990s.)
  4.  Expansion Relations - Extends discourse and develops the narrative.
  *  Conjunction: Arguments share the same relation to a broader situation.
  *  (E.g., I can adjust my insurance or pay a different premium.)
  *  Disjunction: Arguments are alternatives, either or both may hold.
  *  (E.g., We could offer scholarships or provide tax credits.)
  *  Equivalence: Two arguments describe the same situation differently.
  *  (E.g., The pension fund got a bargain-in other words, the real estate is worth more.)
  *  Exception: One argument describes a general rule, the other an exception.
  *  (E.g., All foreign-trading companies are struggling-except some Japanese firms.)
  *  Instantiation: One argument gives examples of the other.
  *  (E.g., Many firms are struggling, such as Givaudan.)
  *  Level-of-Detail: One argument provides a more detailed account of the other.
  *  (E.g., Movies rely on effects over storytelling-essentially, they fear making PG films.)
  *  Manner: One argument describes how the other happens.
  *  (E.g., He lowered costs by lifting production.)
  *  Substitution: One argument presents an alternative after ruling out another.
  *  (E.g., Instead of selling assets, the firm will spin off divisions.)

{_RESPONSE_PLACEHOLDER}

-> In the above paragraph, does "{_MISSING_SPAN}" indicate any relation? If so, answer YES and point out the relation. The relation should be wraped in a markdown code block. If not, please let me know and answer NO. You're only required to answer the level 1 relation. e.g. Temporal, Contingency, Comparison, Expansion.
\end{lstlisting}
}
\noindent Self-reflection, which ensures the missing span is not mentioned in extracted facts.

{\tiny\begin{lstlisting}
I have a list of extracted facts.
{_STATEMENT_LIST_PLACEHOLDER}

Does any of these facts capturing "{_MISSING_SPAN}"?
For example, for a temporal relation, if a fact explicitly mentions a temporal relation like "before", "after", "simultaneously", etc., you should answer YES.
For a contingency relation, if a fact explicitly mentions a contingency relation like "if", "unless", "because", etc., you should answer YES.
If so, point out the fact and answer YES. If not, please let me know and answer NO.
\end{lstlisting}
}

\subsubsection{Filtering FactBench}
\label{apdx:filtering_factbench_prompt}
{\tiny\begin{lstlisting}
A divergent question is a question with no specific answer, but rather exercises one's ability to think broadly about a certain topic. It's designed to encourage a wide range of responses, creative thinking, and exploration of ideas.

Your task is to determine whether a given input is a divergent question. Please note that if the provided input is not a question, you should respond with 'YES'.

Here are some examples:
Question: What are the top 10 most popular programming languages in 2023?
Answer: YES

Question: Example of a lawsuit filed in a court of law for police misconduct
Answer: YES

Question: What are some young soccer talents from lesser known teams?
Answer: YES

Question: can i play cross platform fortnite
Answer: NO

Question: I am so ashamed to be an engineer
Answer: YES

Question: Voltage doesn't kill, Amperage kills.
Answer: YES

Question: Why do we use 3-point belts?
Answer: NO

Question: How was the Super Mario Brothers 3 shortcut found?
Answer: NO

Now it's your turn. Please answer YES or NO to the following questions:
Question: 
\end{lstlisting}}

\subsubsection{Recall of Human-Annotated Fact}
\label{apdx:human_fact_recall}
{\tiny\begin{lstlisting}
Here is a claim and a claim list. Please select the claim from the list that supports the claim. 
If none of the claims support the claim, please say 'None of the above'.
If you need to combine multiple claims to support it, you should aslo say 'None of the above'.
Claim: <claim>
Claim list: <claim_list>
\end{lstlisting}}

\subsubsection{Entailment of Reference Fact and Model Response}
\label{apdx:factual_recall}
{\tiny\begin{lstlisting}
I will provide a claim and a paragraph. Please answer me if the claim is supported by the paragraph. If supported, please say 'Yes'. If not supported or irrelevant, please say 'No'.
Claim: <claim>
Paragraph: <paragraph>
\end{lstlisting}}

\subsubsection{Refining Extracted Facts}
\label{apdx:refining}
We have two prompts to refine the extracted facts. One is for fixing incomplete facts:

{\tiny\begin{lstlisting}
# Task
I will give you a question, an answer and a claim extracted from the answer. However, the claim is problematic. Typically, the problems can be: (1) dependent claims, which means they cannot be correctly understood without the necessary context of the answer; (2) hallucinated claims, which means they are not supported by the answer; (3) self-duplicated claims, which means they repeat information from another part of the text within itself.
I will also tell you the type of the problems and the reason of why this claim is problematic. Please revise the claim to make them self-contained based on this reason. When adding context, use the MINIMAL number of words necessary FROM the answer.
Please shown me the refined claim (DO NOT say anything else. ONLY the refined claim). Wrapped with markdown syntax. i.e.

Example:
# Question
What is the best way to go to Boston from Oakland

# Answer
The best way to travel from Oakland to Boston depends on your budget, time constraints, and personal preferences. One option is to fly from Oakland International Airport (OAK) to Boston Logan International Airport (BOS) with a major airline such as American Airlines, Delta Air Lines, or United Airlines. Flight duration is approximately 5 hours, and you can book a direct or connecting flight depending on your schedule. Another option is to take a train or bus, but this would be a longer journey, taking around 72 hours with multiple changes. Taking a train would involve boarding the Amtrak Coast Starlight from Oakland to Chicago, then transferring to the Lake Shore Limited to Boston, while taking a bus would involve companies like Greyhound or Megabus with multiple transfers. However, flying is generally the fastest and most convenient option.

# Problematic Claim
Taking the train or the bus would be a longer journey.

# Error Type and Reason
dependent: The claim mention "a longer journey", but does not mention the comparison.

# Revised Claim
```
Taking the train or the bus would be a longer journey compared to flying.
```

Your turn:

# Question
<Question>

# Answer
<Answer>

# Problematic Claim
<Claim>

# Error Type and Reason
<Reason>

# Refined Claim
```
\end{lstlisting}}
\noindent The other one is for adding missing fact:

{\tiny\begin{lstlisting}
I will give you a question, an answer and a claim extracted from the answer.
I will give you a span of text that describes a relationship between two events/sentences in the answer. These relationships are either temporal or contingency relationships. Temporal relationships are those that describe the order of events, while contingency relationships are those that describe the cause-effect relationship between events.
I will also tell you the type of relationship. You need to write a sentence that describes the relationship between the two events/sentences in the span, using the minimal number of words possible.

# Question
<Question>

# Answer
<Answer>

# Span
<Span>

# Relationship Type
<RelationshipType>

Please shown me the refined relationship. Wrapped with markdown syntax. For example:
```
<Refined Relationship>
```
\end{lstlisting}}

\subsection{Annotation Interface}
\label{apdx:interface}
\begin{figure*}
    \centering
    \includegraphics[width=1\textwidth]{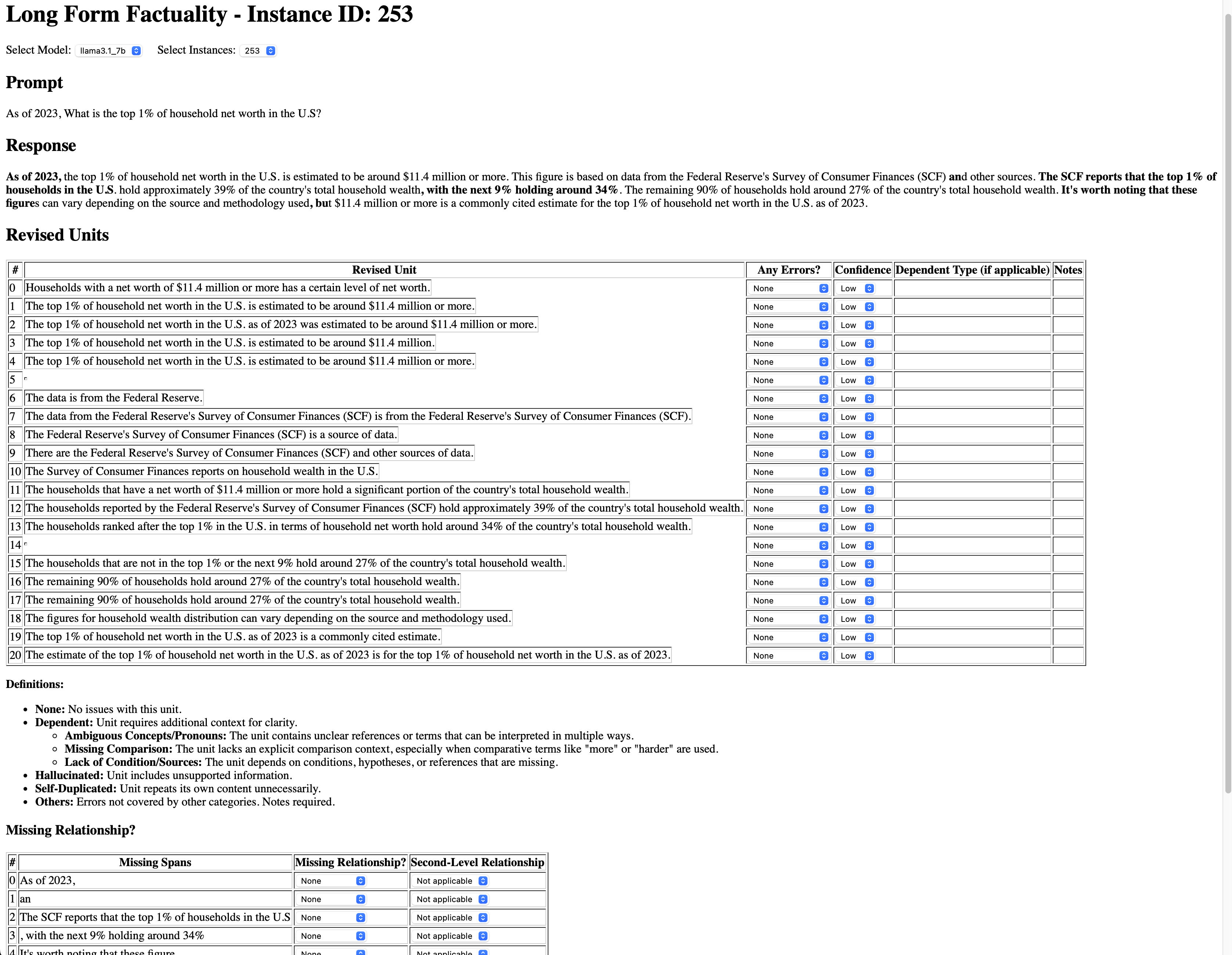}
    \caption{Annotation Interface}
\end{figure*}

\end{document}